\documentclass[11pt, a4paper]{article}
\usepackage[utf8]{inputenc}
\usepackage[T1]{fontenc}
\usepackage{geometry}
\usepackage{graphicx}
\usepackage{booktabs}
\usepackage{hyperref}
\usepackage{amsmath}
\usepackage{amssymb}
\usepackage{authblk}
\usepackage{newtxtext}
\usepackage{newtxmath}
\usepackage{enumitem}

\title{D-CoT: Disciplined Chain-of-Thought Learning for Efficient Reasoning in Small Language Models}

\author{Shunsuke Ubukata}
\affil{Toyo University\\ \texttt{gitpullpull@gmail.com}}

\date{}

\begin{document}

\maketitle

\begin{abstract}
Chain-of-Thought (CoT) distillation from Large Language Models (LLMs) often induces ``overthinking'' in Small Language Models (SLMs), leading to performance degradation and excessive token consumption. In this study, we propose \textbf{Disciplined Chain-of-Thought (D-CoT)}, a novel framework that enforces a structured reasoning process using control tags---such as \texttt{<TEMP\_LOW>} for fact-checking and \texttt{<TEMP\_HIGH>} for multi-perspective exploration---as auxiliary scaffolding during training. By optimizing the CoT trajectory, D-CoT suppresses reasoning drift and simultaneously achieves token reduction and performance improvement. We demonstrate the efficacy of our approach on Qwen3-8B: with only 5,000 training samples, D-CoT significantly boosts accuracy on \textbf{GPQA-diamond by 9.9\%} and \textbf{MMLU-Pro (0-shot) by 9.1\%}, while drastically reducing computational costs. Furthermore, we confirm that the model internalizes this disciplined thought structure, maintaining high performance even without explicit control tags during inference.
\end{abstract}

\section{Introduction}

In recent years, reasoning models represented by DeepSeek-R1 \cite{guo2025deepseek} have acquired the ability to perform complex, multi-step thinking through post-training using extensive Chain-of-Thought (CoT). Consequently, approaches that distill knowledge into Small Language Models (SLMs) using high-quality reasoning processes generated by frontier models are rapidly spreading. 

However, we argue that simply copying and distilling the thought processes of frontier models into SLMs does not necessarily lead to improved reasoning capabilities. Due to limited model capacity, SLMs often fail to control complex contexts, resulting in text drift and unnecessary thought loops---a phenomenon Wu et al.\ \cite{wu2026overthinking} term ``overthinking.'' This leads to a degradation in both token efficiency and accuracy. Existing methods remain limited to passive approaches that retroactively remove CoT segments. We contend that such passive reduction sacrifices exploration diversity and is fundamentally ill-suited for high-difficulty, general-purpose tasks.

To address this limitation, we propose \textbf{Disciplined Chain-of-Thought (D-CoT)}, a reasoning framework optimized for SLMs that reconstructs the order and structure of thought rather than merely removing it. By utilizing control tags (e.g., \texttt{<TEMP\_LOW>}) as scaffolding during training, our method enables the model to learn a disciplined reasoning process.

Our contributions are as follows:
\begin{itemize}[nosep]
    \item We introduce D-CoT, a method to mitigate overthinking in SLMs by structuring the reasoning process via control tags.
    \item We demonstrate that D-CoT achieves substantial performance gains: a 9.9\% improvement on GPQA-diamond \cite{rein2024gpqa} and a 9.1\% improvement on MMLU-Pro \cite{wang2024mmlupro} (0-shot) using only ${\sim}5{,}000$ training samples.
    \item We confirm a significant reduction in average output tokens, verifying that D-CoT optimizes inference efficiency.
    \item We show that the model internalizes the disciplined thought structure, achieving state-of-the-art results even when control tags are not explicitly invoked during inference.
\end{itemize}

\section{Related Work}

\subsection{CoT Distillation and Overthinking}
Wu et al.\ \cite{wu2026overthinking} revealed a scaling law where the relationship between task accuracy and CoT length follows an inverted U-shape. Since the final probability of a correct answer is the product of the success probabilities of each step, SLMs with higher per-step error rates suffer rapid performance degradation when the CoT exceeds an optimal length. While frontier models are trained with Reinforcement Learning (RL) to generate long CoTs for complex tasks, distilling these directly into SLMs causes the length to exceed the SLM's optimal capacity, triggering ``overthinking.''

\subsection{DLCoT and Its Limitations}
As an attempt to eliminate CoT redundancy, Luo et al.\ \cite{luo2025dlcot} proposed DLCoT (Deconstructing Long Chain-of-Thought). This method decomposes CoT into four components and statically filters out redundant paths. However, this approach has two critical limitations. First, its effectiveness is unverified outside of mathematical tasks. Second, Luo et al.'s own experiments showed a significant performance drop on AIME2024 (from 53.3\% to 40.0\%) \cite{luo2025dlcot}. This indicates that simple static data filtering strips away the diversity of exploration required for high-difficulty problems. We argue that static filtering alone is insufficient to maintain an SLM's trial-and-error capabilities while improving efficiency.

\section{Proposed Method: D-CoT}

To prevent overthinking in SLMs, we propose a learning framework that controls the ``mode of thought'' rather than passively decomposing data. D-CoT maintains exploration diversity by using control tags as scaffolding during training, allowing the model to internalize structured thinking.

\subsection{Role of Control Tags}
We introduce three types of control tags to explicitly regulate the nature of the output in each mode. These tags aim to intentionally couple the logicality of the output with the concept of ``Temperature'' acquired during pre-training. Table~\ref{tab:tags} defines the role of each tag.

\begin{table}[h]
\centering
\caption{Definition of Thinking Modes by Control Tags}
\label{tab:tags}
\begin{tabular}{ll}
\toprule
\textbf{Control Tag} & \textbf{Role} \\
\midrule
\texttt{<TEMP\_LOW>} & Fact-checking, listing prerequisites, and constraints. \\
\texttt{<TEMP\_MID>} & Normal response, algorithmic processing, and calculation. \\
\texttt{<TEMP\_HIGH>} & Creative solutions, multi-perspective viewpoints. \\
\bottomrule
\end{tabular}
\end{table}

\begin{figure*}[t]
    \centering
    \includegraphics[width=\textwidth]{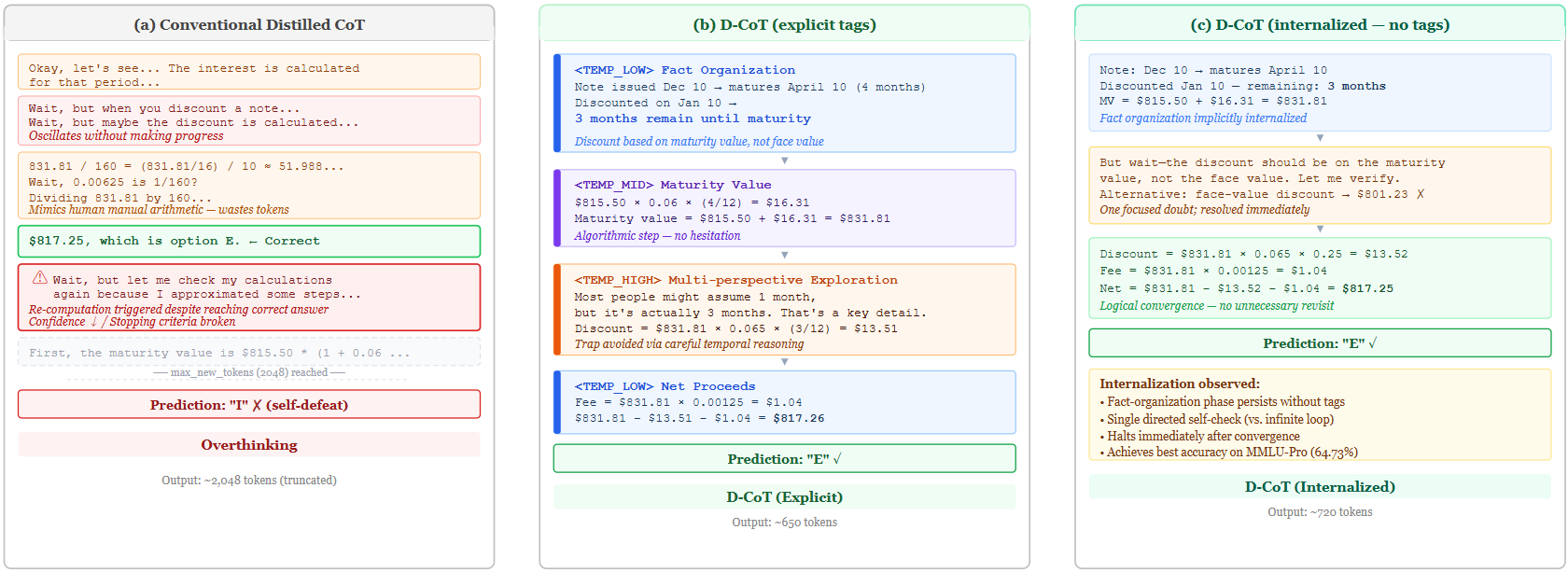}
    \caption{Comparison of reasoning traces on MMLU-Pro \#238 (promissory note discount problem, Qwen3-8B). (a)~Conventional distilled CoT exhibits overthinking: the model reaches the correct answer but continues recalculating, ultimately exhausting the token budget and producing an incorrect prediction. (b)~D-CoT with explicit control tags structures reasoning into fact organization (\texttt{<TEMP\_LOW>}), multi-perspective exploration (\texttt{<TEMP\_HIGH>}), and algorithmic computation (\texttt{<TEMP\_MID>}), converging efficiently. (c)~After D-CoT training, the model internalizes this disciplined structure and maintains organized reasoning \emph{without} explicit tags, achieving the best accuracy on MMLU-Pro (64.73\%).}
    \label{fig:reasoning_comparison}
\end{figure*}

Traditional distilled CoT often exhibits wandering and unnecessary recalculations even after reaching a correct answer. In contrast, D-CoT utilizes tags like \texttt{<TEMP\_LOW>} for initial fact-sorting and \texttt{<TEMP\_HIGH>} to avoid pitfalls through multi-perspective exploration, efficiently converging to the correct answer via \texttt{<TEMP\_MID>}. Figure~\ref{fig:reasoning_comparison} illustrates this contrast on a concrete example.

Importantly, the tag transition order is \emph{not} fixed. Responses may begin without any tag (using default temperature for situational acknowledgment), and the subsequent tag sequence is determined by the nature of the task. For instance, a regulatory compliance problem may require \texttt{<TEMP\_LOW>} first to establish legal constraints, whereas a logistics crisis may begin with dimensional analysis before creative exploration. This flexibility is encoded in the training data via explicit reasoning plans (see Section~\ref{sec:dataset}).

\subsection{Learning via ORPO}
We employ Odds Ratio Preference Optimization (ORPO) \cite{hong2024orpo} instead of standard SFT or DPO. We chose ORPO for two reasons: (1) Preliminary experiments indicated that SFT alone incurred an ``alignment tax'' that degraded the base model's generalization performance. (2) ORPO does not require a reference model and directly integrates the difference between Chosen and Rejected responses into SFT. This allows for more effective learning of the disciplined thinking structure compared to DPO alone.

\subsection{Dataset Construction}
\label{sec:dataset}

\subsubsection{Domain Selection and Contamination Avoidance}
A critical design decision is the \emph{deliberate} use of training domains unrelated to the evaluation benchmarks. In preliminary experiments, we found that generating CoT-style training data with Qwen3-235B on topics similar to MMLU-Pro or GPQA led to significant data contamination, as the teacher model reproduced benchmark-adjacent content. To avoid this and to isolate the effect of structured reasoning from domain-specific knowledge transfer, we selected 7 domains that are rarely represented in academic benchmarks but require persistent, multi-step problem-solving: Legacy IT Operations, Corporate Politics, DIY Engineering, Supply Chain Logistics, Regulatory Compliance, Cybersecurity Incident Response, and Event Crisis Management. From these, 119 scenario topics were defined across 5 instruction templates (emergency response, procedure drafting, crisis management, risk analysis, and constraint optimization).

This domain separation serves a dual purpose: (1)~it ensures that any performance improvement on MMLU-Pro and GPQA can be attributed to the acquisition of \emph{reasoning structure} rather than domain knowledge leakage, and (2)~it enables rigorous decontamination verification, as described in Section~\ref{sec:decontamination}.

\subsubsection{Sample Structure}
Each training sample consists of six fields generated by the teacher model:
\begin{itemize}[nosep]
    \item \textbf{\texttt{user\_prompt}}: A naturalistic, complex query with real-world constraints (e.g., ``we have zero budget,'' ``the server cannot be restarted'').
    \item \textbf{\texttt{thought\_chosen}}: A meta-reasoning plan for the ideal response, explicitly deciding \emph{where} and \emph{why} to use each control tag (e.g., ``Phase 1---Hazard Assessment \texttt{<TEMP\_LOW>}: define the danger zone objectively\ldots\ Token plan: LOW $\to$ MID $\to$ HIGH $\to$ LOW'').
    \item \textbf{\texttt{chosen\_response}}: The ideal response following the plan, with control tags prefixing each segment.
    \item \textbf{\texttt{thought\_rejected}}: A meta-reasoning plan simulating a ``competent but misguided'' persona that leads to a specific error.
    \item \textbf{\texttt{rejected\_response}}: A plausible but flawed response exhibiting the targeted failure mode.
    \item \textbf{\texttt{reasoning}}: A one-sentence summary of the critical flaw for quality assurance (not used during training).
\end{itemize}

During ORPO training, the model receives the \texttt{user\_prompt} as input, with \texttt{thought\_chosen} + \texttt{chosen\_response} as the preferred output and \texttt{thought\_rejected} + \texttt{rejected\_response} as the dispreferred output. The \texttt{thought} fields are placed inside \texttt{<think>} blocks, following Qwen3's native thinking format \cite{yang2025qwen3} in which the model performs internal reasoning within dedicated \texttt{<think>}\ldots\texttt{</think>} delimiters before producing a visible response. This means the model learns not only to produce structured responses, but also to \emph{plan} which control tags to use and why---a key mechanism underlying the internalization phenomenon.

\subsubsection{Rejection Category Design}
We defined 31 rejection categories organized along three axes to ensure the model learns nuanced preference signals:
\begin{itemize}[nosep]
    \item \textbf{Tag usage failures}: Complete Token Absence (no tags used), Token Omission (missing necessary tag transitions), Premature Creativity (using \texttt{<TEMP\_HIGH>} before establishing facts with \texttt{<TEMP\_LOW>}).
    \item \textbf{Content quality failures}: Architectural Naivety, Latency Blindness, Compliance Hallucination---where the response is structurally well-formed but contains domain-specific errors.
    \item \textbf{Safety and judgment failures}: Safety Violation (e.g., routing people near live power lines), Dangerous Minimization (understating legal or physical risks).
\end{itemize}
Critically, rejected responses are designed to be ``competent but misguided''---coherent and well-written, but with a specific fatal flaw. This forces the model to learn the \emph{alignment between tag usage and content quality}, rather than simply distinguishing good writing from bad.

\subsubsection{Generation and Decontamination}
\label{sec:decontamination}
We used Qwen3-235B-Instruct (A22B) via OpenRouter API as the teacher model. Using Few-Shot System Prompts with three fully worked examples, the model generated JSON files containing the six-field structure described above. A total of 5,181 samples were produced via random sampling from the 18,445 possible combinations of scenarios, templates, and rejection categories.

\begin{figure*}[t]
    \centering
    \begin{minipage}[t]{0.48\textwidth}
        \centering
        \includegraphics[width=\textwidth]{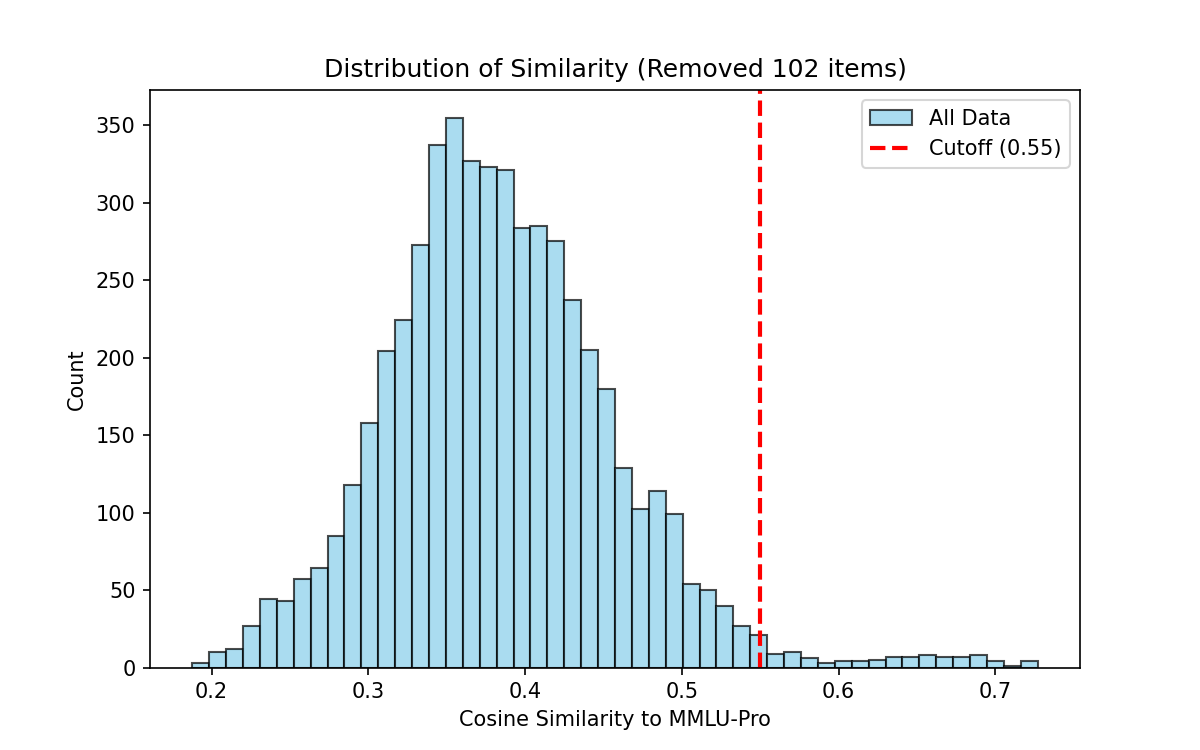}
    \end{minipage}
    \hfill
    \begin{minipage}[t]{0.48\textwidth}
        \centering
        \includegraphics[width=\textwidth]{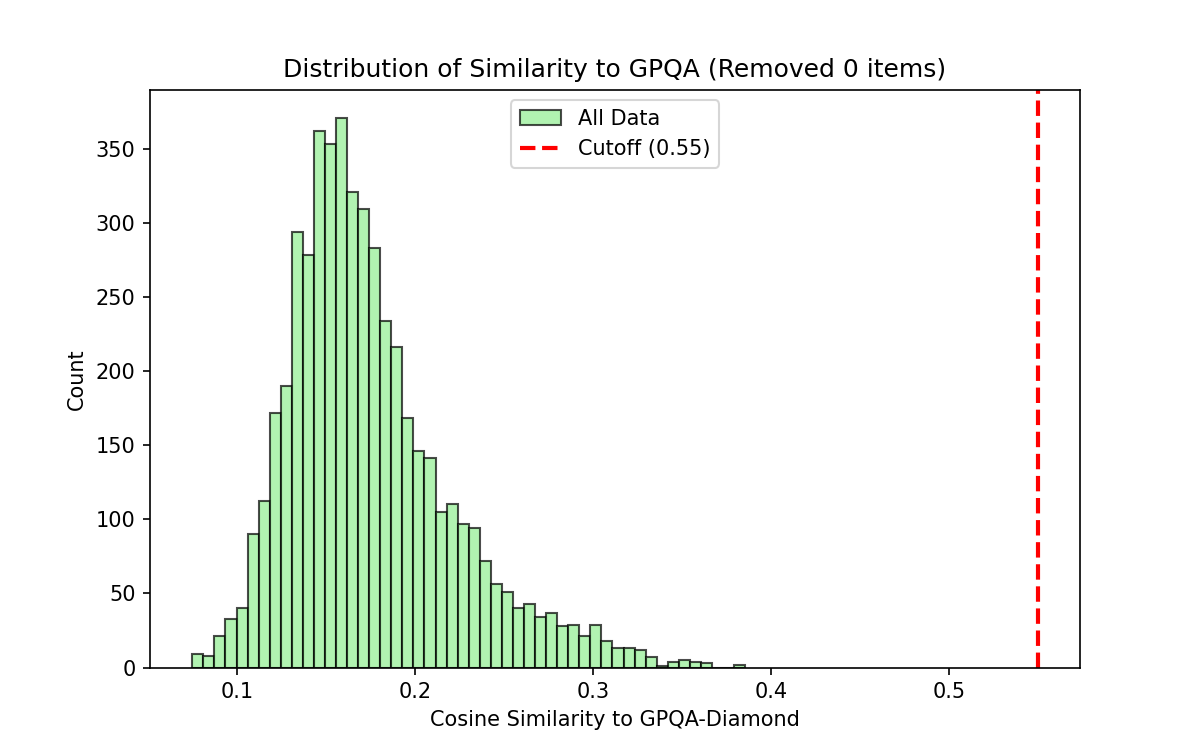}
    \end{minipage}
    \caption{Cosine similarity distributions between D-CoT training samples and benchmark test sets. Left: MMLU-Pro (102 samples removed above the 0.55 threshold). Right: GPQA-Diamond (0 samples removed). The low similarity confirms that training domains are well-separated from evaluation benchmarks, and that observed performance gains stem from reasoning structure acquisition rather than knowledge leakage.}
    \label{fig:contamination}
\end{figure*}

To rigorously prevent benchmark contamination, we applied dual-criterion filtering against the GPQA and MMLU-Pro test sets: samples with Cosine Similarity $> 0.55$ (using \texttt{all-mpnet-base-v2} embeddings) \emph{or} 13-gram overlap were removed. Figure~\ref{fig:contamination} shows the similarity distributions. Against MMLU-Pro, the distribution peaks around 0.35 with 102 samples removed above the threshold. Against GPQA-diamond, the distribution peaks around 0.15 with \emph{zero} samples removed, confirming the extreme domain separation between our training data and the science-focused GPQA benchmark. The final clean dataset contains 5,079 samples.

\subsection{Training Setup}
We used \textbf{Qwen3-8B} \cite{yang2025qwen3} as the base model, selected for its balance of performance, efficiency, and native CoT capability. After decontamination, 5,079 Chosen/Rejected pairs were used for training. LoRA training was conducted on an RTX 5090 using the Unsloth library. The optimizer was Lion (8-bit) with a learning rate of 4e-6, $\beta_{\text{ORPO}} = 0.1$, batch size of 1 (gradient accumulation = 8), warmup ratio of 0.1, cosine learning rate scheduler, bfloat16 precision, and 2 epochs.

\section{Inference and Benchmarks}

\subsection{Evaluation Tasks}
Since standard benchmarks (MMLU, GSM8K) are saturated for modern SLMs, we adopted high-difficulty benchmarks to quantitatively detect qualitative changes in the reasoning process:
\begin{itemize}[nosep]
    \item \textbf{MMLU-Pro:} Requires multi-step reasoning with 10 choices. Evaluated 0-shot on all 12k questions.
    \item \textbf{GPQA-diamond:} Extremely difficult science/expert questions (198 questions). Evaluated 0-shot (5-seed average) to assess pure reasoning ability without prompt engineering bias.
\end{itemize}

\subsection{Experimental Setup}
We compared six conditions combining the model (Base vs.\ D-CoT), temperature settings (Locked vs.\ Dynamic), and prompts (Base vs.\ Custom).
For ``Dynamic'' settings, we implemented a mechanism to switch sampling temperature in real-time based on the generated tag: 0.3 for \texttt{<TEMP\_LOW>}, 0.6 for \texttt{<TEMP\_MID>}, and 0.8 for \texttt{<TEMP\_HIGH>}. The ``Locked'' setting used a fixed temperature of 0.6. Following Qwen3's official recommendations \cite{yang2025qwen3}, we set Top-P $= 0.95$ and Top-K $= 20$; greedy decoding (temperature $= 0$) was avoided as it induces infinite loops in Qwen3. The maximum output length was 2,048 tokens for MMLU-Pro and 8,196 for GPQA-diamond.

\section{Experimental Results}

\subsection{MMLU-Pro Results}
Table~\ref{tab:results} summarizes the results. On MMLU-Pro, the Base model achieved 55.66\% accuracy with 1,742 average tokens. D-CoT (LoRA) with the Base Prompt achieved \textbf{64.73\%}, a substantial improvement of 9.07 points. Notably, the highest accuracy was recorded \emph{without} explicitly forcing the Custom Prompt, confirming the internalization of the reasoning process. D-CoT also reduced the average token count to 1,199 (Custom Prompt), a \textbf{31.2\% reduction} compared to the baseline, successfully eliminating overthinking.

\subsection{GPQA-diamond Results}
On GPQA-diamond, D-CoT (Dynamic/Custom) achieved \textbf{52.93\%} accuracy, outperforming the Base model (43.03\%) by \textbf{9.9 points}. The average token count was reduced from 5,875 (Base) to 2,073 (D-CoT), a massive \textbf{64.7\% reduction}. The ``Null rate''---the proportion of responses from which no valid answer could be extracted via regex---dropped from 30.91\% (Base) to less than 5\% (D-CoT), indicating that the model acquired the ability to reach valid conclusions through disciplined reasoning. Even after applying Null-corrected scores (where Null responses are assigned random-chance credit of 25\%, corresponding to the 4-choice format of GPQA), D-CoT (54.47\%) still surpasses the corrected Base score (50.76\%), confirming that the accuracy gains are not merely an artifact of reduced Null rates.

\begin{table}[ht]
\centering
\caption{Comparison of Accuracy, Null Rate, and Average Output Tokens on Qwen3-8B}
\label{tab:results}
\resizebox{\textwidth}{!}{%
\begin{tabular}{lllcccccc}
\toprule
\textbf{Method} & \textbf{Temp.} & \textbf{Prompt} & \textbf{MMLU-Pro Acc.} & \textbf{MMLU-Pro Tokens} & \textbf{GPQA Acc.} & \textbf{GPQA Null Rate} & \textbf{GPQA Null-Corr.} & \textbf{GPQA Tokens} \\
\midrule
Base & Locked & Base & 55.66\% & 1,742 & 43.03\% & 30.91\% & 50.76\% & 5,875 \\
Base & Locked & Custom & 58.24\% & 1,595 & 45.05\% & 24.44\% & 51.16\% & 5,539 \\
Base & Dynamic & Custom & 57.64\% & 1,600 & 45.05\% & 24.44\% & 51.16\% & 5,539 \\
\midrule
\textbf{D-CoT (LoRA)} & Locked & Base & \textbf{64.73\%} & 1,496 & 52.82\% & 4.24\% & 53.88\% & 2,772 \\
\textbf{D-CoT (LoRA)} & Locked & Custom & 62.92\% & \textbf{1,199} & 51.41\% & \textbf{1.82\%} & 53.27\% & \textbf{2,073} \\
\textbf{D-CoT (LoRA)} & Dynamic & Custom & 63.43\% & 1,202 & \textbf{52.93\%} & 3.54\% & \textbf{54.47\%} & 2,153 \\
\bottomrule
\end{tabular}%
}
\end{table}

\subsection{Pareto Improvement}
\begin{figure*}[t]
    \centering
    \includegraphics[width=\textwidth]{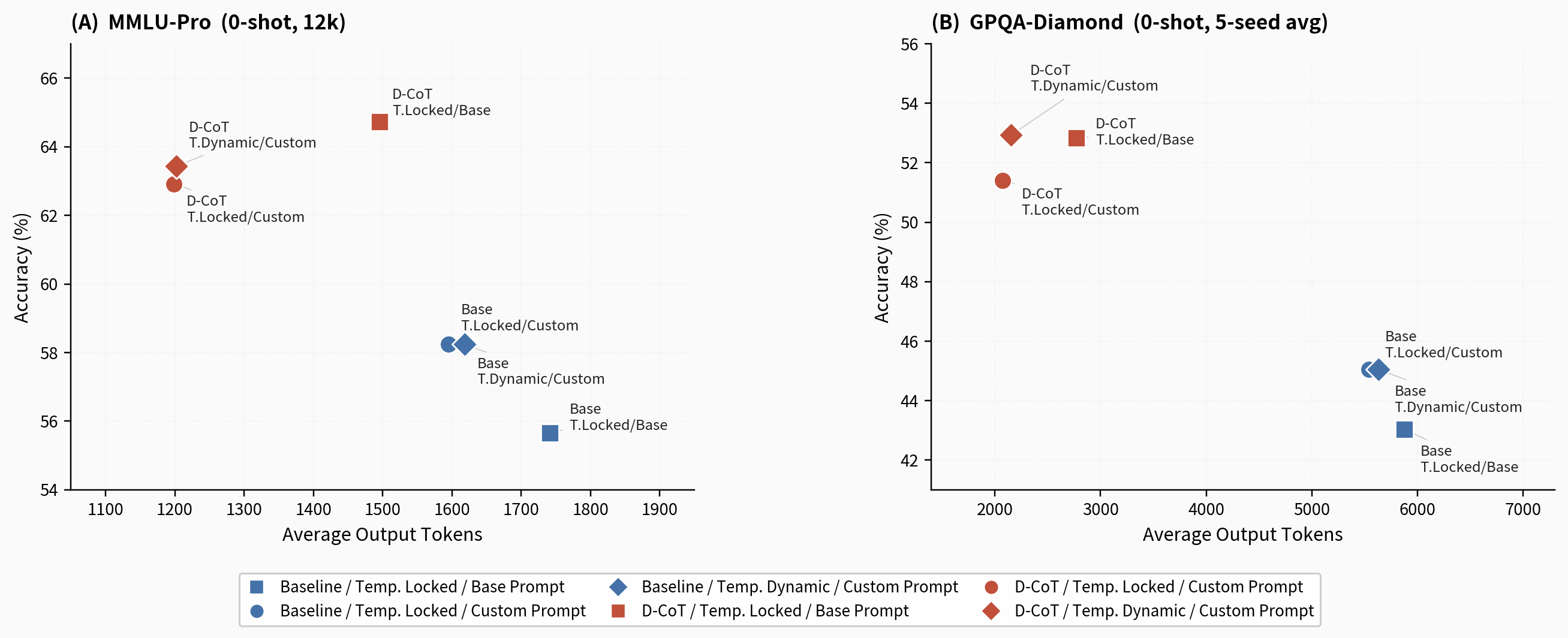}
    \caption{D-CoT: Accuracy vs.\ Average Output Tokens (Qwen3-8B, 0-shot). (A)~MMLU-Pro (12k questions). (B)~GPQA-Diamond (5-seed average). D-CoT conditions (red) cluster in the upper-left (high accuracy, low tokens), demonstrating a Pareto improvement over the Baseline conditions (blue). The inter-group separation far exceeds variation from temperature or prompt settings.}
    \label{fig:scatter}
\end{figure*}

Figure~\ref{fig:scatter} visualizes the accuracy--token trade-off across all six conditions. D-CoT samples are concentrated in the high-accuracy, low-token quadrant (upper-left), forming a clear Pareto frontier. This demonstrates that D-CoT does not trade off accuracy for brevity but simultaneously improves both. The separation between the Base and D-CoT clusters is substantially larger than the variation caused by temperature or prompt settings, confirming that structural optimization is the dominant factor.

\section{Discussion}

\subsection{Internalization of Disciplined Thought}

The most striking finding is that the highest MMLU-Pro accuracy (64.73\%) was achieved \emph{without} explicit control tags during inference. As illustrated in Figure~\ref{fig:reasoning_comparison}(c), the internalized model exhibits three characteristic behaviors that distinguish it from the Base model's overthinking:

\paragraph{Implicit fact organization.} The model begins with structured premise enumeration (``Note: Dec 10 $\to$ matures April 10 / Discounted Jan 10 $\to$ remaining: 3 months'') without any \texttt{<TEMP\_LOW>} tag. This mirrors the fact-sorting phase learned during training.

\paragraph{Single directed self-check vs.\ infinite loops.} When encountering an ambiguous point, the Base model enters ``Wait, but\ldots'' loops that consume hundreds of tokens without progress---a pattern we term \emph{probabilistic oscillation}, where the model picks up low-probability alternative branches rather than committing to a calculation path. In contrast, the D-CoT model performs a single, focused verification (``But wait---the discount should be on the maturity value, not the face value. Let me verify.'') and resolves it immediately. This behavioral shift from undirected hesitation to directed self-correction is the hallmark of internalized discipline. We observe that this probabilistic oscillation is pervasive across both benchmarks, appearing in financial mathematics (MMLU-Pro) and science reasoning (GPQA) alike.

\paragraph{Suppression of unnecessary human-mimicking arithmetic.} The Base model wastes hundreds of tokens simulating step-by-step manual arithmetic (e.g., ``831.81 / 160 = (831.81/16) / 10 $\approx$ 51.988\ldots''), reproducing the kind of longhand calculation processes prevalent in pre-training data. The D-CoT model skips these unnecessary simulations and proceeds directly to the result. Notably, we did \emph{not} explicitly train the model to avoid such behavior; it emerged as a side effect of learning structured reasoning, suggesting that disciplined thought organization naturally suppresses less efficient reasoning patterns.

\paragraph{Why tagless inference outperforms explicit tags.}
We hypothesize two complementary mechanisms: (1)~Outputting control tags consumes inference resources (tokens and attention capacity) that could otherwise be allocated to reasoning. Internalizing the structure eliminates this overhead. (2)~Explicit tags impose forced mode-switching at predetermined boundaries, which may occasionally interrupt the model's natural reasoning flow. Without tags, the model transitions between reasoning modes at organically optimal points.

\subsection{Cross-Domain Generalization}

A natural question is why training on domains such as cybersecurity incident response and supply chain logistics improves performance on academic science questions (GPQA) and multi-domain reasoning (MMLU-Pro). As shown in Figure~\ref{fig:contamination}, the training data has near-zero semantic overlap with GPQA (peak similarity ${\approx}\,0.15$) and low overlap with MMLU-Pro (peak similarity ${\approx}\,0.35$, with 102 borderline samples removed).

We argue that D-CoT teaches \emph{reasoning structure}, not domain knowledge. The control tags encode domain-independent cognitive operations: gathering constraints before exploring solutions (\texttt{<TEMP\_LOW>} $\to$ \texttt{<TEMP\_HIGH>}), performing focused computation (\texttt{<TEMP\_MID>}), and converging to conclusions. These operations transfer across domains because overthinking is itself a domain-independent pathology---the probabilistic oscillation and post-answer recalculation observed in Figure~\ref{fig:reasoning_comparison}(a) occur regardless of whether the task is financial mathematics or organic chemistry.

Furthermore, the deliberate use of unrelated training domains provides a methodological advantage: any observed improvement can be attributed to structural reasoning gains rather than knowledge leakage, offering a cleaner experimental signal than training on in-domain data.

\subsection{Efficacy of Dynamic Temperature}
The improvement from dynamic temperature control was marginal (0.51--1.52 points). Considering the implementation complexity, the structural optimization provided by D-CoT is the dominant factor in performance gain. This suggests that the control tags' primary value lies in organizing the \emph{training signal}, not in controlling sampling behavior at inference time.

\section{Conclusion}

We proposed D-CoT, a framework for learning disciplined reasoning processes in SLMs. Our approach achieved simultaneous improvements in accuracy and token efficiency on high-difficulty benchmarks, effectively mitigating ``overthinking.'' The model successfully internalized the structured reasoning pattern, achieving the best performance without explicit control tags. Future work will verify the effectiveness of this method on other model families and scales, and analyze attention maps to understand the internal mechanisms of internalization.

\section*{Acknowledgments}
We utilized Anthropic Claude Sonnet 4.5 and Google Gemini 3 Pro for literature review and code implementation assistance. All generated content was verified and revised by the author.

\section*{Reproducibility}
Implementation code (ORPO training scripts, evaluation scripts), benchmark results (GPQA-diamond, MMLU-Pro), and the full dataset (5,181 Chosen/Rejected pairs with control tags) are publicly available.\footnote{Code: \url{https://github.com/gitpullpull/DisciplinedChainOfThought}}\footnote{Benchmarks: \url{https://huggingface.co/datasets/gitpullpull/D-CoT-Benchmarks}}\footnote{Dataset: \url{https://huggingface.co/datasets/gitpullpull/D-CoT-datasets}}

\bibliographystyle{alpha}

\end{document}